\documentclass[letterpaper, 10 pt, conference]{ieeeconf}  % Comment this line out if you need a4paper

\IEEEoverridecommandlockouts                              % This command is only needed if 
\usepackage{booktabs}
\usepackage{setspace}
\usepackage{cite}
\usepackage{amsmath,amssymb,amsfonts}
\usepackage{algorithmic}
\usepackage{graphicx}
\usepackage{textcomp}
\usepackage{xcolor}
\usepackage{multirow}
\usepackage{bm}
\usepackage{amsmath}
\usepackage{booktabs}
\usepackage{microtype}
\usepackage{bbm}
\usepackage{hyperref}

\title{
DexDribbler: Learning Dexterous Soccer Manipulation via Dynamic Supervision
}

\author{Yutong Hu*, Kehan Wen and Fisher Yu% <-this % stops a space
\thanks{The authors are with VIS Group at ETH Zurich, 8092 Zurich, Switzerland.}%(email: huyuto@ethz.ch; kehwen@student.ethz.ch; i@yf.io).}%
\thanks{$^{*}$Corresponding Author: huyuto@ethz.ch}%
}

\begin{document}

\maketitle{}
\thispagestyle{empty}
\pagestyle{empty}

%%%%%%%%%%%%%%%%%%%%%%%%%%%%%%%%%%%%%%%%%%%%%%%%%%%%%%%%%%%%%%%%%%%%%%%%%%%%%%%%
\begin{abstract}

% As a case study demonstrating the need for robots to execute complex behavioral strategies in dynamic environments, robot soccer has been a longstanding grand challenge for AI and robotics, since at least the formation of the RoboCup competition. While there have already been robots that can dribble a ball in the wild thanks to advances in deep reinforcement learning, the ball can't yet be controlled on a seemingly easier flat smooth floor. The complexity arises because a ball does not stop spontaneously on such a surface, requiring the robot to constantly adjust to the ball's movement and strategically position itself to redirect it. This macro-regulated behavior does not emerge through the existing trial-and-error process, even with added entropy gains in high-dimensional action spaces. We incorporate a feedback control block into the reinforcement learning training pipeline to provide a dynamic supervision to guide the policy network in learning such behavior. We further implement an improved ball dynamic model, an extended context-aided estimator, and a comprehensive ball observer to facilitate the policy transfer to the real world. Building on these advancements, we propose the first learning-based solutions that enable soccer robots to perform sophisticated maneuvers like sharp cuts and turns on flat surfaces, enhancing their potential to play soccer under official rules and prescribed field conditions. Video and code are available at https://website.with.videos.

Learning dexterous locomotion policy for legged robots is becoming increasingly popular due to its ability to handle diverse terrains and resemble intelligent behaviors. However, joint manipulation of moving objects and locomotion with legs, such as playing soccer, receive scant attention in the learning community, although it is natural for humans and smart animals. A key challenge to solve this multitask problem is to infer the objectives of locomotion from the states and targets of the manipulated objects. The implicit relation between the object states and robot locomotion can be hard to capture directly from the training experience. We propose adding a feedback control block to compute the necessary body-level movement accurately and using the outputs as dynamic joint-level locomotion supervision explicitly. We further utilize an improved ball dynamic model, an extended context-aided estimator, and a comprehensive ball observer to facilitate transferring policy learned in simulation to the real world. We observe that our learning scheme can not only make the policy network converge faster but also enable soccer robots to perform sophisticated maneuvers like sharp cuts and turns on flat surfaces, a capability that was lacking in previous methods. Video and code are available at \href{https://github.com/SysCV/soccer-player}{github.com/SysCV/soccer-player}.

\end{abstract}

%%%%%%%%%%%%%%%%%%%%%%%%%%%%%%%%%%%%%%%%%%%%%%%%%%%%%%%%%%%%%%%%%%%%%%%%%%%%%%%%
\section{INTRODUCTION}

Artificial Intelligence is getting embedded into robotic bodies to become more accessible and useful. Among those body designs, legged locomotion stands out for its flexibility and anthropomorphic characteristics. Thanks to the advances in control theories and engineering, legged robots \cite{jiDribbleBotDynamicLegged2023, jenelten2024dtc, margolisWalkTheseWays2022} nowadays can perform complex maneuvers like walking, running, dancing, crawling, jumping, etc., in pre-determined environments, much richer than other forms of locomotion such as wheels. Those technologies further enable the creation of humanoid robots\cite{hwangboLearningAgileDynamic2019a}, poised to revolutionize social productivity.

\begin{figure}[t]
    \centering
    \includegraphics[width=0.5\textwidth]{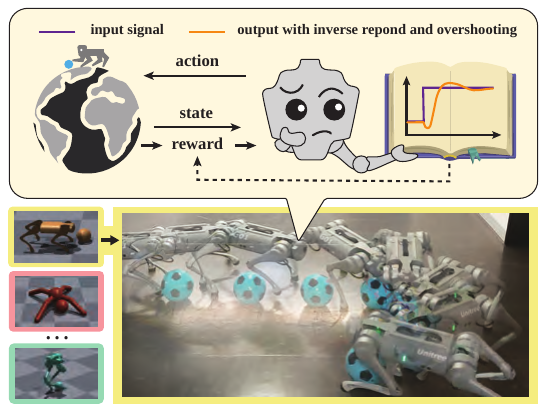}
    \caption{\textbf{Demonstration of a DexDribbler}. Guided by a feedback controller, the robot learns pinpoint coordination between body movement and feet motion in simulator. This enables it to execute ``deliberate overshooting'' --- a critical technique for performing sharp cuts and turns while dribbling on flat and smooth surfaces in real world.}
    \label{fig:big-turn}
\end{figure}

However, the legs can do much more than locomotion. They can also manipulate objects as well. Humans constantly move objects with their legs for work and entertainment, such as playing soccer. Similar to manipulation with arms and hands, the legs will also need to cope with the uncertainties of the interacted objects and the environment. More challenging than the arms, the legs must perform those manipulations while supporting the body. This difficulty is compounded by challenging terrains and high-degree freedom of articulated limbs, prohibiting effective solutions based on traditional control.

Recently, reinforcement learning has enjoyed great success in learning generalizable legged locomotion policy on various terrains \cite{margolisWalkTheseWays2022, rudinLearningWalkMinutes2022, nahrendraDreamWaQLearningRobust2023, kumarRMARapidMotor2021, hwangboLearningAgileDynamic2019a, jiConcurrentTrainingControl2022b}. However, despite the prevalence of legged manipulation in the natural world, few works focus on the challenging legged manipulation problem. Several learning-based methods based on model-free reinforcement learning, such as DribbleBot\cite{jiDribbleBotDynamicLegged2023} and OP3\cite{haarnojaLearningAgileSoccer2023}  were proposed for this task. Those policy learning try to infer limb actions from the states and objectives of manipulated objects. Unfortunately, this indirect supervision usually leads to poor performance and generalizability in the manipulation task. For example, in the soccer case, the learned policy cannot dribble the ball on smooth surfaces as the ball speed can be high, and the overrun of the robot is necessary to stop the ball.

This paper aims to design an efficient and effective learning scheme for the legged manipulation task. We mainly consider the case of ball manipulation because it is common in human world and the ball motion is complex. We observe that although it is challenging to infer the limb articulation on unknown terrains from the ball status, it is much easier to estimate the necessary body motion based on traditional control. Therefore, we propose integrating a feedback block based on PID control to estimate the body motion and supervise the policy training. This supervision is dynamic because it depends on the target states. Besides, a neural-aided Kalman Filter is used to estimate the ball states more accurately and further facilitate the complex dynamic manipulation of the ball in real-world deployment. 

Our main experiments are conducted on quadrupedal dribbler for robot soccer.
Besides the improved results in quantitative evaluations, we find that our learning policy can perform sharp cuts and turns on flat and smooth surfaces. Our method is not only effective in the case of dynamic ball dribbling. In scenarios where the dimension of an agent's low-level action space is large, it can offer a way to integrate high-level priors into the learning process. It results in a globally more optimal policy while preserving the accuracy of low-level maneuvers. We also test our methods on multiple types of quadrupedal and bipedal robots and various terrains in simulator, to test the generalizability and adaptability of our method. Our dynamic supervision can be easily adapted to those scenarios and the resulting models perform significantly better than the state-of-the-art.

\section{Related Work}
% The soccer dribbling problem using quadrupedal robots can be viewed as intersection of three research areas in robotics: dynamic object manipulation, locomotion control for legged robotics, and the robot soccer.

\subsection{Dynamic Object Manipulation} Reaction to an externally moving object, e.g. catching a moving ball, is the most studied form of dynamic manipulation for robot. Such tasks require the collaboration of perception, prediction, planning and control. Solutions vary from cases where fully observed object trajectories using motion capture systems \cite{dongCatchBallAccurate2020a} to partially observed ones by ego-cameras \cite{suCatchingFlyingBall2017}. Successful demonstrations are showcased on a variety of robotics platforms ranging from robotic arms \cite{kimCatchingObjectsFlight2014a}, legged robots \cite{forraiEventbasedAgileObject2023} to drones \cite{suCatchingFlyingBall2017}.

Subsequent exploration delves into cases where the dynamic motion of the object comes from the repeated internal imposition given by robots, such as blindly juggling balls with open-loop controlled robotic arm \cite{ploegerHighAccelerationReinforcement2020}, learning a residue physics model to throw objects into a box \cite{zengTossingBotLearningThrow2020}, continuously pushing a block following a given path \cite{heinsMobileManipulationUnknown2021}, and using reinforcement learning for coordinated manipulation while legs and arms are moving at the same time \cite{fuDeepWholeBodyControl2022}.
% When the object being dynamically manipulated is a ball, the tasks are often related to robot soccer, which we will discuss separately.
However, when the manipulator is the leg instead of the arm and hand, the manipulation needs to achieve both supporting the body and actuating the targets, which leads to a multi-task policy problem.

% Quadrupedal robots, with their capability to effectively traverse uneven terrains, are increasingly being utilized in both indoor and outdoor environments. Bipedal robots, though not as flexible as the former because of greater instability and controlling complexity, are still able to demonstrate human-like walking ability \cite{liBridgingModelbasedSafety2022a}.

\subsection{Legged Locomotion}  A classical approach to control quadrupedal and bipedal robots is using a model-based controller \cite{bledtMITCheetahDesign2018} to plan the joint motion while minimizing the tracking error and energy consumption. Recently, deep reinforcement learning methods \cite{margolisWalkTheseWays2022, rudinLearningWalkMinutes2022, nahrendraDreamWaQLearningRobust2023, kumarRMARapidMotor2021, hwangboLearningAgileDynamic2019a, jiConcurrentTrainingControl2022b} showed a great capability in complex and agile maneuvers without compromising real-time performance. With a carefully designed neural context estimator \cite{nahrendraDreamWaQLearningRobust2023, kumarRMARapidMotor2021, hwangboLearningAgileDynamic2019a, jiConcurrentTrainingControl2022b}, the robot can traverse multiple kind of terrain blindly with a single learned policy. Utilizing ego-centric or exteroceptive visual information \cite{yangLearningVisionGuidedQuadrupedal2022, loquercioLearningVisualLocomotion2022, leeLearningQuadrupedalLocomotion2020} can further improve the performance. Further, model-based approaches can be combined with learning-based methods, as a system identification wrapper for safety guarantee during navigation \cite{liBridgingModelbasedSafety2022a}, as a back-mounted block making policy output more accurately implemented by joint actuators \cite{lyuCompositeControlStrategy}, or as intermediate block to provide feed-forward signal to reduce the delay and error \cite{zhouDeepNeuralNetworks2020a}. Recently, DTC \cite{jenelten2024dtc} employs guidance from traditional controllers to enhance learned policy's precision and robustness, a concept we also embrace. However, unlike DTC's reliance on a traditional trajectory planner during deployment, our method has engraved motion references only during training, thus eliminating the need for runtime guidance.

\subsection{Legs as Manipulators: Robot Soccer}
Achieving human-level soccer skills with robots remains an enduring goal in the robotics community~\cite{kitanoRoboCupRobotWorld1997}. But most presented soccer skills by legged robots, such as kicking \cite{velosoPlayingSoccerLegged1998} and goalkeeping \cite{friedmannVersatileHighqualityMotions2008}, use rule-based motion primitives due to the complex dynamics. Recently, by leveraging deep reinforcement learning, quadrupedal robots demonstrate the capacity to perform multiple skills separately such as dribbling a ball to a target \cite{bohezImitateRepurposeLearning2022a} on grass and continuously dribbling a ball on multiple rough terrain \cite{jiDribbleBotDynamicLegged2023}. Kicker \cite{jiHierarchicalReinforcementLearning2022} and Goal Keeper \cite{huangCreatingDynamicQuadrupedal2022} quadruped robots can also be trained to precisely shoot a soccer ball to a target or jump to block a shoot. A series of works \cite{haarnojaLearningAgileSoccer2023, byravanNeRF2RealSim2realTransfer2022b, liuMotorControlTeam2022} successfully transfer a soccer play skills from bipedal players in simulators to simplified yard in real world. The agent can combine kicking, fall-recovering and intercepting skills to score a goal, yet can not actively keep the ball within the playing area. The demands for robots to continuously adapt to the ball's movement and strategically position themselves for redirection inspire us to propose our approach: incorporating body-level moving guidance into a learning based joint-level control policy, rather than doing policy optimization purely replying on carefully designed reward \cite{jiDribbleBotDynamicLegged2023, haarnojaLearningAgileSoccer2023}.

\section{Overview}

We introduce a novel framework for teaching robots to perform dynamic ball manipulation tasks efficiently and effectively. Sec.\ref{sec-sim} presents the training phase in simulator: we take vectorized robot’s pro-prioceptive results and ball’s position as state observation, and using manufacturer-provided URDF model to simulate action execution result. Apart from rewards, explicit body motion guidance is calculated in a feedback formulation that takes the ball's and robot's current and target states at each timestep to direct the training of our policy network. In addtion, domain randomization are applied jointly with a context-aided estimator network to improve learning efficiency and sim-to-real transfer quality.

%To make robots efficiently and effectively learn to perform dynamic ball manipulation, our approach calculates guidance for limb articulation in a feedback formulation, taking ball's and robot's current and target states at each timestep during simulation, directing the training of our policy network. After training, the motion guidance was exclusively embeded to the policy network. Therefore, we can zero-shot transfer the neural policy from real to real without any further motion guidance or adjustments on network. Only an additional measurement module is required to provide ball position from robot's ego-observation, giving the same form of state input to get action output from learned policy network.

%\fy{This should be moved to experiment section.}

%\fy{Need a real overview paragraph here to follow the introduction.}

%During the training phase, all the physical parameters and states are accessible. We take vectorized robot's proprioceptive results and ball's position as observation, and using manufacturer-provided URDF model to simulate action execution results. To improve learning efficiency and sim-to-real transfer quality, domain randomization are applied jointly with a context-aided estimator network, as well as our newly proposed feedback guidance module, which will be detailed in \ref{pid-section}.

Upon completing training in simulator, limb articulation policy is fully embedded within the neural network, enabling a zero-shot sim-to-real transfer without the need for further motion reference or policy adjustments. A crucial component for real-world deployment is an additional measurement module that provides the ball's position from the robot's ego-perception (two onboard 210$^{\circ}$ field-of-view fisheye cameras, one facing forward and one facing downward), ensuring state input consistent with how the policy is trained. Sec.\ref{sec-real} introduces the deployment phase, including how to get the ball state estimation by combining the detected bounding boxes, the neural estimator, geometric prior and dynamic models. Sec.\ref{sec-exp} gives evaluation results in both simulated and real-world environments showing effectiveness and overall improvements from our method.

%During the real world deployment, the same observation vector can be provide from the onboard IMU, motor encoders and joysticks, expect the ball position vector in robot body frame. The ball position vector is obtained from image captured by two onboard 210$^{\circ}$ field-of-view fisheye cameras, one facing forward and one facing downward. And the final position is given by a Kalman filter combining the detected bounding boxes, the neural estimator, geometric prior and dynamic models.

\section{Training Phase in Simulator} \label{sec-sim}

\begin{figure}[t]
    \centering
    \includegraphics[width=0.5\textwidth]{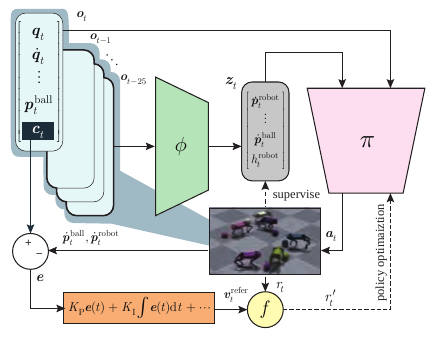}
    \caption{\textbf{Training pipeline during learning phase in simulator.} Ground truth states can be obtained from simulator, and are used both for estimator network supervision, and for body speed feed-back computation.}
    \label{fig:sim}
\end{figure}

\subsection{Environment Design} \label{ball-drag-sec}
\subsubsection{State Definition}

The whole policy network $\Pi$ contains two separate blocks: the context estimator $\phi$ and the actor $\pi$. The input to the whole network $\Pi$ is a observation set $[\bm{o}_t,\bm{o}_{t-1} \cdots,\bm{o}_{t-24}]$ consisting of the 25-step history of robot joint positions $\bm{q}$ and velocities $\bm{\dot q}$, ball position $\bm{p}_t$, gravity unit vector in the body frame $\bm{g}_t$, global body yaw $\psi_t$, and timing reference variables $\tau_t$ as defined in \cite{margolisWalkTheseWays2022}. The commands $\bm{c}_t$ consist of the target ball velocities $v^{cmd}_x , v^{cmd}_y$ in the global frame. The action space $\bm{a}_t$ is the target position of the twelve joints $\bm{q}^{cmd}_t$, which will be taken as input a low-level PD controller with $\bm{k}_p = 20.0, \bm{k}_d = 0.5$, adjusting the power of joint motor to perform given movement in the simulator and real world.

\subsubsection{Ball-Terrain Interaction Model} 

The drag force on the ball as it rolls on the ground is different from the sliding friction force on the robot's foot as it comes into contact with the ground. Different from DribbleBot's approach to assume drag force proportional to the square of the ball velocity which is similar to aerodynamic drag, we follow the Relative Velocity Dependent (RVD) rolling friction model \cite{zhouRollingFrictionDynamic1999} where rolling friction is directly proportional to the relative angular velocity. In our case, if the ball is pure-rolling, then the relative angular velocity is directly proportional to the velocity of the ball. If the ball is not rolling purely, then the ground will be sufficiently smooth that there will be very little friction. Therefore, we can further approximate the drag effect as
\begin{equation}
    {{\ddot{\bm{p}}_{ball}}} =  \frac{\bm{F}_{drag}}{m} = \frac{-C_{drag}}{m} \cdot \dot{\bm{p}}_{ball} = - C_{D} \cdot \dot{\bm{p}}_{ball} .
\end{equation}

During the simulation, we perform domain randomization to use different values of $C_D$ within the range $[-0.1,0.5]$ to emulate terrains with various resistance forces, such as a field with tall grass (high), pavement (low), wooden floor (close to zero) and possible inclination and unevenness (temporary negative). 

Considering the ball alone as a dynamical system, the state-space equation of the system is

\begin{equation}
\frac{\mathrm{{d}}}{{\mathrm{{d}}t}} {\begin{bmatrix}
\bm{p}\\
{\dot{\bm{p}}}
\end{bmatrix}} = {\begin{bmatrix}
0 & 1\\
0 & -C_D
\end{bmatrix}}{\begin{bmatrix}
\bm{p}\\
{\dot{\bm{p}}}
\end{bmatrix}},
\end{equation}
where $C_D$ determines the matrix's Eigen Value. When $C_D > 0 $, the system is stable and the ball will always stop, so the robot just needs not to fall down while waiting for the ball to stop. However, for the cases of asymptotic stable and unstable when $C_D \leq 0$ , the ball can not stop itself, the robot needs to make the right move at the right time in order to maintain control of the ball. This explains why although a rough terrain seems challenging for locomotion, controlling the ball on a smooth terrain turns out to be more difficult.

\subsubsection{Context Definition}

% Please add the following required packages to your document preamble:
% \usepackage{booktabs}
% \usepackage{graphicx}
\begin{table}[]
\centering
\caption{parameters for domain randomization}
\label{tab:my-table}
\begin{tabular}{@{}ccc@{}}
\toprule
Dynamics   Parameter                & Range             & Units \\ \midrule
Payload Mass                        & {[}-1.0,2.0{]}    & kg    \\
Motor Damping                      & {[}90,110{]}      & \%    \\
Motor Stiffness                     & {[}95,105{]}      & \%    \\
Joint Calibration                   & {[}-0.02,0.02{]}  & rad   \\
Robot-Terrain Friction              & {[}0.10,2.00{]}   & -     \\
Robot-Terrain Restitution           & {[}0.00,2.00{]}   & -     \\
Robot Center of Mass   Displacement & {[}-0.15,0.15{]}  & m     \\
Mass                                & {[}0.20,0.40{]} & kg    \\
Camera Frame Arrival Rate           & {[}0.3,0.7{]}     & -     \\
Teleporting Position                & {[}0.0,1.0{]}     & m     \\
Perturbation Velocity               & {[}0.0,0.3{]}     & m/s  \\
Ball-Terrain Drag Coefficient       & {[}-0.1,0.5{]}     & -     \\
$v_x^{\text{cmd}}$       & {[}-1.5,1.5{]}    & m/s  \\
$v_y^{\text{cmd}}$       & {[}-1.5,1.5{]}    & m/s  \\ \bottomrule
\end{tabular}%
\end{table}

Prior works \cite{kumarRMARapidMotor2021,hwangboLearningAgileDynamic2019a,nahrendraDreamWaQLearningRobust2023,jiConcurrentTrainingControl2022b} found some contextual information that is not accessible from robot's onboard sensors can be inferred explicitly or implicitly from historical action-observation responds. Assisting the network with those estimated context can be useful for sim-to-real transfer. Motivated by those prior works, we build a context-aided estimator network $\phi$ to better adapt the environment. We focus on three parts of the value: i) The environment parameter that changes during domain randomization, as in Table.\ref{tab:my-table}.
ii) The robot body linear and angular velocity, and the body height that can't be obtained from on-board sensors. iii) The ball's true relative position and velocity, as well as the prediction of its position in the coming two time step, as a partial understanding of the ball's forward dynamic. 

All the values are estimated in a explicit way $\bm{z}_t = \phi(\bm{O}_t)$ because some of the value can be useful for other module mentioned in \ref{kalman}. And $\phi$ can be learned supervised by the ground-truth states $\bm{s}_t$ obtained from the simulator. %Although the absolute numerical value of the property parameters like friction or ball-drag coefficient don't hold the same meaning in the real world domain, they can still reflect the correspondence between those properties in the real world and those in the simulator. Therefore %
Getting actions from a network conditioned on those estimated context $\bm{a}_{t+1} = \pi(\bm{O}_t, \bm{z}_t) $ can better cope with sim-to-real gap.

\subsubsection{Reward Design}

Our reward function closely follows the original DribbletBot \cite{jiDribbleBotDynamicLegged2023}, to highlight the effect of the modules added in the pipeline instead of reward tuning. The whole reward function is consisted of: i) a Task reward for tracking the commanded ball velocity in the global reference frame. ii) Auxiliary terms encouraging the robot to be close to the ball to make interaction between the robot and the ball possible. And reward when the robot is directly facing the ball, to promote ball visibility in the camera. iii) A set of gait reward terms and standard safety reward terms to induce a well-regulated gait pattern, penalize dangerous angular joint and facilitate the sim-to-real transfer. 

\subsection{Feedback Controller in the Learning Loop} \label{pid-section}
 Behavior to bring the ball to a stop on a rough terrain was more of the robot waiting for the ball to decelerate, rather than actively stopping the ball with the feet. However, to actively stop the ball or make the ball heading the inverse direction on a smooth terrain, robot needs to surpass the ball by a short distance first, providing proper position and enough time to take action to stop the ball from continuing its movement --- an ``overshoot'' that would sacrificing the short-term reward, as the robot accelerates in the opposite direction from $\bm{c}_t$, but allows for greater cumulative returns.

The existing RL framework should have cope with such problem theoretically. But for this dribbling case study, the behavior failed to emerged by itself. This required some careful design of a guidance mechanism, to make the policy informed of the higher-body-level suggestion whereas the action space controlled by the policy is the lower-motor-level angle, leading the way robot controlling its own body aligned with our expecting behavior.

We note that the traditional feedback controller such as a PID controller can generate manageable overshoots. And particularly in non-minimum-phase systems, there are situations where the initial response to a control input is in the opposite direction of the desired outcome. This phenomenon, which is typically referred to as ``inverse response'', is inevitable in some systems and a necessary process to achieve the final goal. Inspired by this, we integrate a feedback block in the framework of RL to generate such behavior and to guide the learning of the policy, as shown in Fig.\ref{fig:sim},
\begin{equation}
\begin{gathered}
\bm{v}_t^{\text {refer }}=K_{\text{P}}\left(\dot{\bm{p}}_t^{\text {ball }}-\dot{\bm{p}}_t^{\text {robot }}\right)+K_\text{I} \int\left(\dot{\bm{p}}_t^{\text{ball }}-\dot{\bm{p}}_t^{\text {robot }}\right) \\
+K_{\text{CMD}}\left(\dot{\bm{p}}_t^{\text {ball }}-\bm{c}_t \right),
\end{gathered}
\end{equation}
Where $K_\text{P} = 0.5, K_\text{I} = 4.0, K_\text{CMD} = 1.0$, and all other variables used for computing is obtainable from states $\bm{s}_t$. 
Having the reference speed given by the feedback controller. We need to integrate such guidance into the policy to make the it able to generate low level joint motion to meet the requirement of the high-level guidance. Here we shape the total reward with regard of how well current action align to the high-level speed guidance.
\begin{equation}
r_t^{\prime} = f(r_t,\bm{s}_t) = e^{-\Delta{p}_t^{\text {feet}} / \sigma}\left(r_t + e ^{-\left|\bm{v}_t-\bm{v}_t^{\text{refer}}\right|}\right),
\end{equation}
where $\sigma = 0.02$ and 
\begin{equation}\label{eq:delta_feet}
    \Delta p_t^{\text {feet }}=\sum_{i=1}^4\left(1-\mathbbm{1}_i^{\text {near }}\right) \cdot \left|\bm{p}_t^{\text {foot-}i}-\bm{p}^{\text {foot-}i}_{\text{RH}}(\bm{v}_t^{\text{refer }})\right|.
\end{equation}

 In Eq.\eqref{eq:delta_feet}, the suggested kinematic motion of every foot $\bm{p}^{\text {foot-}i}_{\text{RH}}(\bm{v}_t^{\text{refer }})$ is calculated from the traditional Raibert Heuristic Gait Generater \cite{raibertLeggedRobotsThat1986} given the body velocity and time reference, and $\mathbbm{1}_i^{\text {near }}$ is an indicator function which becomes 1 once the distance of the ball relative to the $i^{\text{th}}$ foot below a predefined threshold of 10 cm. It guarantees that when the ball is close to a particular leg, the flexibility of agile manuer is not limited.

\subsection{Neural Network Architecture and Optimization:}
We use Proximal Policy Optimization (PPO) \cite{schulmanProximalPolicyOptimization2017} in an asymmetric way to train our soccer dribbling policy. the policy (actor) receives temporal partial observations $\bm{O}_t$ and the context vector $\bm{z}_t$ estimated from context network detached from the gradient propagation, while the value network (critic) receives the full state $\bm{s}_t$. The actor and critic are separate neural networks having three fully-connected hidden layers of the same sizes [512, 256, 128]. The context network is a neural network having two fully-connected hidden layers of sizes [512, 256], and the parameters are trained according to the MSE loss between the estimate states, with the corresponding ground truth state abtained from the simulator. As the three networks are independent of each other, the parameter sets can be jointly optimized within one gradient descent step. We use ELU as activation function for all networks.

\section{Deployment Phase in Real World}\label{sec-real}

During the real world deployment, the whole control pipeline can be transferred zero-shot. As the network parameters no longer require update, the policy network $\pi$ and estimator network $\phi$ are set to inference mode with no grad propagation, and the critic network can be dropped. Most of the element in the observation vector $\bm{o}_t$ can be provide from the onboard IMU and motor encoders. Command $\bm{c}_t$ can be read from joysticks for real time control or from a predefined script for reproducible benchmark. We choose to command the robot in the local body frame of the robot at the first time step, which makes the same command direction constant and easy to interpret from the recording. The global orientation is given by the yaw $\psi_t$ obtained from the 9-DOF IMU. Due to random magnet disturbances and accumulated shift, absolute positional drift is unavoidable, but short-term direction consistency can be guaranteed.

The only element in the observation vector that can not be obtained directly from on-board sensors is the three dimensional ball position $\bm{p}^{\text{ball}}_t$. What we can get from the onboard cameras are RGB pixel tensors $\bm I$. Therefore, a ball detector is required to get the position of the ball in the image. Then the detection measurement needs to be further converted to vectorized position in robot body frame --- the same formulation as we get in the simulation phase.

\subsection{Detection network for Ball Perception}

In our project, we used a specialized ball-detection system using the YOLOv8 object detection model\cite{tervenComprehensiveReviewYOLO2023}, initially trained on the Coco\cite{linMicrosoftCOCOCommon2014} image dataset. Due to the wide-angle fisheye lens of the camera, and drastic changes in occlusion and shading when the ball is at different positions around the body, off-the-shelf models trained on standard rectified internet images faced challenges in accurately detecting the spherical soccer ball, especially when it is at the edges of the field of view --- making it easily distracted and occluded by legs. To overcome this issue, we finetune the YOLOv8 model with a dataset of 1000 manually annotated fisheye images of soccer balls captured from our robot's perspective. This dataset included various scenarios, containing images where the ball was positioned at the periphery of the frame and right under the body. We further enhance the network's robustness by employing standard image augmentation techniques, including horizontal and vertical flipping, HSV value shifting and blur effects. Finally, we are able to get detection boxes $ \bm B=\text{YOLO}(\bm I)=[x_\text{min},y_\text{min},x_\text{max},y_\text{max}]^{\text{T}} $  in pixel space, having 0.948 mAP@0.5.
\subsection{Fusion of Perception result} \label{kalman}

In the equidistant fisheye model, the distance between a pixel in the image and the principal point is directly proportional to the angle of incidence:
$ r = f \cdot \theta$.

\begin{figure}[t]
    \centering
    \includegraphics[width=0.5\textwidth]{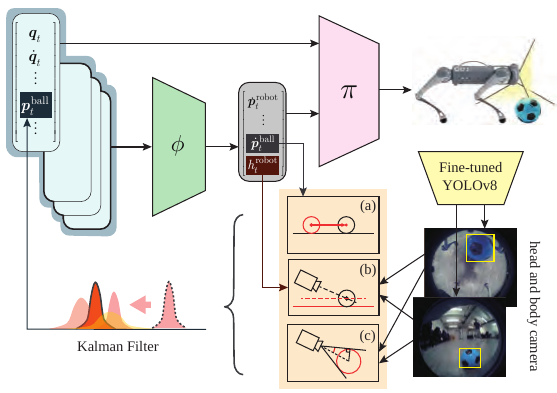}
    \caption{\textbf{Deployment pipeline in real world.} The ball position vector is calculated by a kalman filter combining: (a) Constant Velocity Model (b) Projection-Intersection Model (c) Viewing Angle Model}
    \label{fig:real}
\end{figure}

\subsubsection{Viewing Angle Model}
Given the bounding box of the ball in pixel coordinates, we first compute the approximate ball diameter in pixel by calculating the Geometric Mean of the box edge $\Delta d = \text{sqrt}\left( {(x_\text{max}-x_\text{min})\cdot(y_\text{min}-y_\text{min})}\right )$.  This pixel distance corresponds to the angle formed between the two sides of the ball and the camera center in the world frame $\Delta\theta=\Delta d/f$, as in Fig.\ref{fig:real}(c). Further, knowing the ball size $D_\text{ball} = 18\text{cm}$ in the real world, we can calculate the distance between the ball and the camera center $D = D_{\text{ball}}/2\sin(\Delta \theta/2)$. With distance scale fixed, the three dimensional position vector of the ball in robot frame can be calculated.

\subsubsection{Projection-Intersection Model}
The center pixel of the ball's bounding box $P_\text{x} = ({x_\text{max}-x_\text{min}})/{2},P_\text{y} = {(y_\text{min}-y_\text{min}})/{2}$ is corresponding to a ray that starts from the camera center having an angle $\theta = \text{sqrt}(P_\text{x}^2 + P_\text{y}^2) / f $ with camera optical axis, and an angle $\alpha = \text{arctan}(P_{\text{x}}/P_{\text{y}})$ with the camera horizon axis. Knowing that ball only moves on the ground and ignoring the swing during the locomotion, we can get the ball's center by calculating the intersection point between the center pixel project ray and the plane of a known height, as in Fig.\ref{fig:real}(b).

\subsubsection{Constant Velocity Model}
As the context estimator also output the estimated speed of the ball $\dot{ \bm{p}}_{t}^{\text{ball}}$, we can calculated the position of the ball in the coming time step simply applying a  constant velocity model $\bm{p}_{t}^{\text{ball}} =  \bm{p}_{t-1}^{\text{ball}} + {\dot{\bm{p}}}^{\text{ball}}_{t-1}\Delta t$. By defining the ball state maintained by the filter (different with the ``state'' taken by the policy) as $\bm{x} = [\bm{p},\dot{\bm{p}} ]^{\mathrm{T}} $. We can formulate a vectorized representation of the dynamical model:
\begin{equation}
{{\bm{x}}_t} = {\begin{bmatrix}

{{{\bm{I}}_{2 \times 2}}}&{{{\bm{I}}_{2 \times 2}} \cdot \Delta t}\\
O&{{{\bm{I}}_{2 \times 2}}}
\end{bmatrix}}  {\begin{bmatrix}
{{{\bm{p}}_{t - 1}}}\\
{{{{\bm{\dot p}}}_{t - 1}}}\end{bmatrix}}  + {\bm{w}} = {\bm{F}} \cdot {{\bm{x}}_{t - 1}} + {\bm{w}},
\end{equation}
where $\bm{w}\sim\mathcal{N}(\bm{O},\bm{Q})$ accounts for the uncertainty of the dynamic model, we set $\bm{Q}_{4 \times 4} = \mathrm{diag}(
0.01,0.01,0.2,0.2)$ 

\subsubsection{Kalman filter combining three models}

Having four position measurements from two different cameras using two separate observation models, and one velocity measurement from the context estimation network. We can formulate the measurement process as 
${\bm{m}} = [\bm{p}_{\text{angle}}^{\text{cam1}},\bm{p}_{\text{angle}}^{\text{cam2}},\bm{p}_{\text{center}}^{\text{cam1}},\bm{p}_{\text{center}}^{\text{cam2}},\dot {\bm{p}}_{\phi}]^{\mathrm{T}} = {\bm{H}\bm{x}} + {\bm{\vartheta}}$, where
\[
{\bm{H}} = \begin{bmatrix}
\bm{I}_{2 \times 2} & \bm{I}_{2 \times 2} & \bm{I}_{2 \times 2} & \bm{I}_{2 \times 2} & \bm{O}_{2 \times 2} \\
\bm{O}_{2 \times 2} & \bm{O}_{2 \times 2} & \bm{O}_{2 \times 2} & \bm{O}_{2 \times 2} & \bm{I}_{2 \times 2}
\end{bmatrix}^{\mathrm{T}}
\]
is the measurement matrix and $\bm{\vartheta}\sim\mathcal{N}(\bm{O},\bm{R})$ is the measurement noise,  we set $\bm{R}_{10 \times 10} = \mathrm{diag}(
0.01,$ $0.01, \cdots 0.01,0.1,0.1)$. Considering we need the ball state before the context estimator having output, we calculated the initial ball position using only the \textit{Viewing Angle Model} and the detection box with higher confidence from two images. After setting the initial ball speed as zero and $\bm{P}_0 = 0.01 \cdot \boldsymbol{I}_{4 \times 4}$, we can calculate ball state within each step using Kalman update procedure:
\begin{equation}
\begin{array}{c}
{{{\bm{\tilde x}}}_t} = {\bm{F}}{{{\bm{\hat x}}}_{t - 1}},\qquad
{{{\bm{\tilde P}}}_t} = {\bm{F}}{{\bm{P}}_{t - 1}}{{\bm{F}}^T} + {\bm{Q}}\\
{{\bm{K}}_t} = {{{\bm{\tilde P}}}_t}{{\bm{H}}^T}{\left( {{\bm{H}}{{{\bm{\tilde P}}}_t}{{\bm{H}}^T} + {\bm{R}}} \right)^{ - 1}}\\
{{{\bm{\hat x}}}_t} = {{{\bm{\tilde x}}}_t} + {{\bm{K}}_t}\left( {{{\bm{z}}_t} - {\bm{H}}{{{\bm{\tilde x}}}_t}} \right),\ {{\bm{P}}_t} = \left( {I - {{\bm{K}}_t}{\bm{H}}} \right){{{\bm{\tilde P}}}_t}
\end{array}    
\end{equation}

Notice that all matrixes are partitionable and that velocity estimation $\dot{\bm{p}}_\phi$ is always available. When some perceptual sources are not available because of missed detection, out of view or occlusion, we can simply disable the corresponding block in matrix calculation. Therefore, the proposed Neural aided Kalman filter is a practical and convenient approach for merging multi-source estimations. However, most measurement processes are simplified to a unit mapping. Further improvements could be made by accurately representing the measurement process and the uncertainty of neural estimators.

\section{Experiments}\label{sec-exp}

For a comparative evaluation for solving the dribbling task as a dynamic object manipulation problem, we compared the following algorithms with access to proprioceptions only:

\textbf{1) DribbleBot}\cite{jiDribbleBotDynamicLegged2023}: The policy was optimized only through roll-out returns, but during domain randomization we extend the ball-drag coefficient below zero to make sure the agent has been exposed to such (smooth and unstable terrain) cases.

\textbf{2) DribbleBot+:} We extend \textbf{DribbleBot}'s existing context estimator network to output all estimate-able random parameters, as an enhancement. Other parts remains the same. 

\textbf{3) DexDribbler:} Our proposed method. The network architecture is exactly the same as \textbf{DribbleBot+}, but during the trainning phase the reward is further shaped by reference body movement generate by the feedback controller.

All the methods above were trained using the same actor and critic network architecture, same domain randomization range and fixed the initial random seeds. 
\subsection{Simulation Performance}
\subsubsection{Learning Performance}

\begin{figure}[t]
    \centering
    \includegraphics[width=0.5\textwidth]{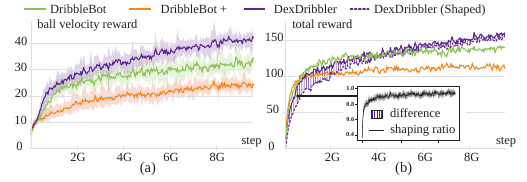}
    \caption{\textbf{Reward curve during training phase.} (a) Task related reward term (b) Summation of all reward terms}
    \label{fig:learn_curve}
\end{figure}

\begin{figure}[t]
    \centering
    \includegraphics[width=0.5\textwidth]{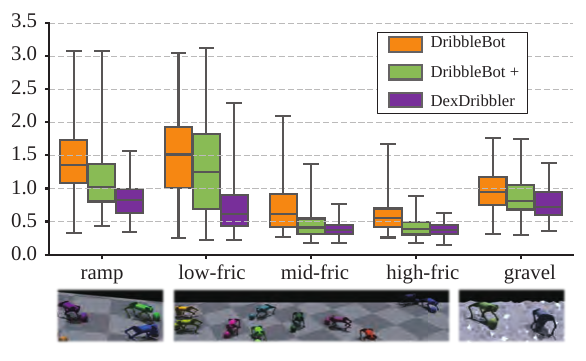}
    \caption{\textbf{Absolute ball velocity tracking error on different terrain.}}
    \label{fig:sim_box}
\end{figure}

We used the Isaac Gym simulator [36] based on a open-source implementation of PPO [12] to synchronously train the policy, value, and estimator networks. We trained 8k domain-randomized agents in parallel. All training was conducted on a single NVIDIA TITAN XP GPU. The expected return from the critic network stabilized after 2 billion global steps, yet the reward curve continued to grow slowly with ongoing training. We documented the training curve over 10 billion global steps, equivalent to approximately 48 hours, to align with the baseline settings (7  billion global steps) and ensure that our improvements were not temporary throughout the whole training phase.

As in Fig.\ref{fig:learn_curve}, our method always obtains the largest reward regarding the ball velocity error  within the same number of iteration steps, meaning that the agent can control the ball the most precisely in the environment compared to others. As for the total reward, we plot the reward before and after guidance shaping. Our method does not get the maximum total reward at the beginning, but as the body movement created by our behavior gradually matches the movement supervision from the feed-back controller, the agent can find a overall more optimal policy, and makes total reward eventually exceeds that of the other methods.

\subsubsection{Ball Controllability on Different Terrain}
We evaluate the command tracking performance in the same simulation environment to evaluate the final performance in with-in distribution cases. The ball-drag coefficient $C_D$ is sample between [-0.1,0.1), [0.1,0.3) and [0.3,0.5] to simulate low, mid and high drag terrain respectively. We add two more environments: a 3$^{\circ}$ ramp and a gravel environment to simulate the out-of-distribution cases. The robot was given random commands for 40 seconds, and the commands were uniformly sampled from [-1.0m/s, 1.0m/s] every ten seconds. For fair comparison, random commands were generated using the same random seed for each policy. Robot controlled by each policy runs 1000 times with different random seeds to verify repeatability. We measured Absolute Tracking Error (ATE) as the performance metric and constructed as shown in Fig.\ref{fig:sim_box}.  Our method consistently outperforms the baselines. Moreover, We can see that as the ball-drag coefficient increases, several methods can perform dribbling more accurately and the difference between them reduces. This confirms our analysis mentioned in Sec. \ref{ball-drag-sec} that high-drag terrain actually makes the task easier.

\subsubsection{Trajectory Following Test}

\begin{figure}[t]
    \centering
    \includegraphics[width=0.5\textwidth]{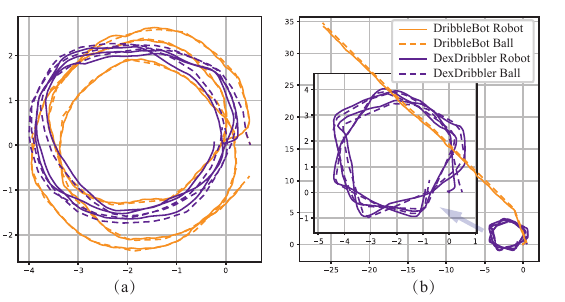}
    \caption{\textbf{Circular trajectory following test in simulator.} (a) middle-drag terrain where $C_D=0.2$ (b) low-drag terrain where $C_D=0$}
    \label{fig:sim_traj}
\end{figure}

We generate the velocity commands to follow a circle trajectory with diameter of 5 meters. Notice that the commands are applied in velocity space, and in position space the system can be considered open-loop, so drift always exists. Nevertheless, the size of the drift error in position space can reflect the following accuracy in velocity space. On high-drag terrain as shown in Fig.\ref{fig:sim_traj}(a), both method can enable robot successfully follow the trajectory, but our method's trajectory has larger overlapping part. However, on low-drag terrain, the baseline method lose the control of the ball. As a result, the ball keeps moving in one direction. Our method, although becoming less accurate, still keeps the ball rolling within a circular trajectory.

\subsection{Real World Performance}
We use the Unitree Go1 robot\footnote{https://www.unitree.com/cn/go1/} and a size 3 soccer ball for all realworld experiments. We zero-shot transfer the policy learned in simulator to the real world, as metioned in Sec.\ref{kalman}. While Dribbling in the real world, the locomotion policy must adapt when the terrain causes the feet to slip or stumble. To keep the ball within control, it additionally needs to do running and kicking adjustment depending on how the ball interacts with the terrain, E.g., on grass, high drag tends to slow down the rolling ball; on smooth floor, low drag may cause it to speed away from the robot; on gravel, the ball changes direction unpredictably as it impacts the terrain surface. 
\subsubsection{Quantitative Results on Diverse Terrains}
\begin{table}[]
\centering
\caption{Real-world dribbling performance evaluation.}
\label{tab:real-table}
\centering
    \includegraphics[width=0.5\textwidth]{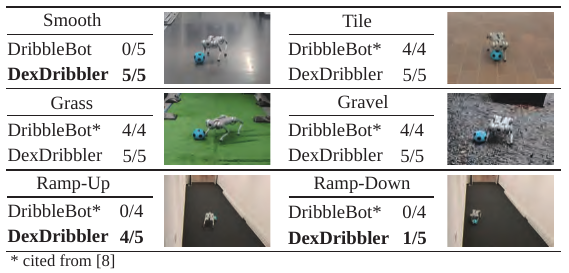}
\end{table}

We quantitatively evaluate the success rate of the fully autonomous behavior of the dribbling policies while executing a scripted trajectory across diverse terrains: the set of [Tile, Grass, Gravel] terrains that Dribblebot already have been tested on, a [Slope] case that have not been solved, and a [Smooth floor] case that is not included in their test. The robot is commanded with a predetermined trajectory: dribble forward at 1.0 m/s for 5 s, then stop the ball. As we don't integrate a auto-recovery maneuver, the failure cases can be easily judged: if the distance between robot and ball exceed 0.5m, or the robot falls. This is a slightly different testing method than the Dribblebot's ``turn-back'' testing. We did this because the ``stopping state'' is crucial, as switching to any dribbling direction is easy when the ball stops, and the ability to stop the ball reflects the overall ball controllability. Additionally, dribbling uphill and downhill are two cases, each presenting their unique challenges, thus seperating them into two dribble-and-stop process allow for a more fine-grained evaluation of performance.

As in Table.\ref{tab:real-table}, in scenarios where Dribblebot has been tested, our method performed the same as they did. On the smooth terrain, Dribblebot can only follow the ball and run behind it, but cannot make it stop. Our method can make a quick cut-off to bring the ball to stop. When dribbling up-hill a ramp, our method can respond quickly to ball shift, blocking the ball from rolling back to its starting point. However, when dribbling down-hill, our method often pitches forward and fall with face when trying to intercept a ball that is accelerating and rolling down. We find that covering random gravity direction during training can address this issue, but for a fair comparison we still report the results of zero-shot deployments under the uniform setting.

\subsubsection{Trajectory Following Test}
\begin{figure}[t]
    \centering
    \includegraphics[width=0.5\textwidth]{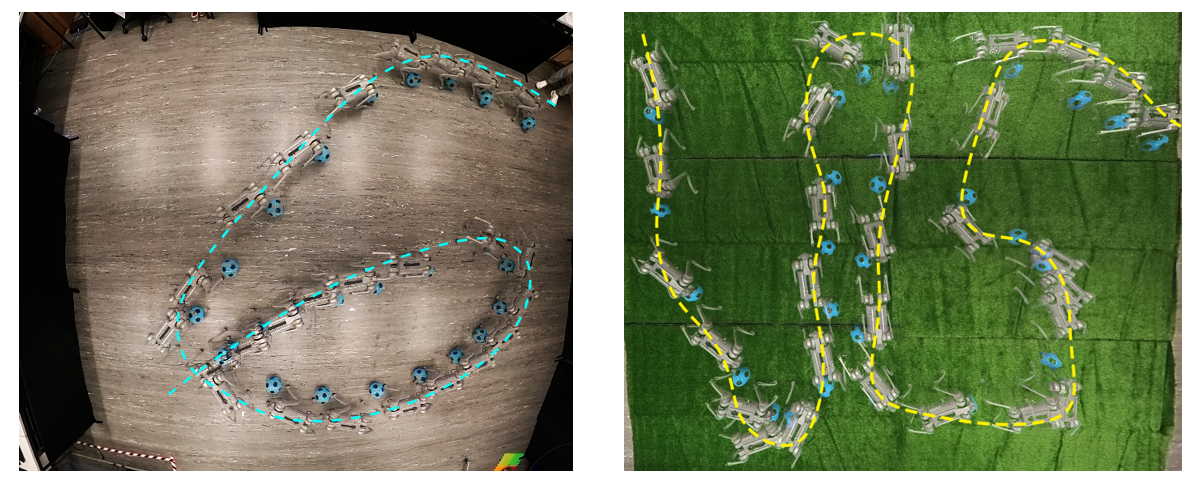}
    \caption{\textbf{Real-world trajectory following test.} The robot is controlled under teleportation to create some recognizable trajectory, ``6'' on smooth terrain and ``VIS'' on grass terrain.}
    \label{fig:real_traj}
\end{figure}
We qualitatively evaluate our dribbling controller under teleoperation on diverse terrains with different ball-terrain dynamics. Because our system operates without a tether or external sensing, long-term safety could not be guaranteed when following only open-loop speed command. Therefore, we make robot receive commands from a joystick manipulate by human as a close-loop spotter. Yet, the policy still needs to be accurate enough and make agile large-angle turns at corners to create recognizable trajectory graphics, especially on smooth and uneven surfaces. Fig.\ref{fig:real_traj} shows stitched overhead photos to illustrate the real-world dribbling performance.

\subsection{Generalizability on similar tasks}

By simply adapting the action dimension and specifying the foot index number, our training pipeline and the original Dribblebot's approach are both naturally compatible for ball dribbling tasks across different legged robot configurations. When we replaced the Unitree Go1 robot with Cassie, a popular large-size bipedal robot, and NAO, a smaller bipedal robot used in the RoboCup Standard Platform League, our method demonstrated a 39.2\% and 11.6\% higher final task-return than Dribblebot's, respectively, after 10 billion training steps in a simulator. These results highlight that the improvements from our method are not limited to a single robot model and show our method's potential for training soccer robots that are ready to compete under the official rules and field conditions of RoboCup.

\section{Conclusion and Future works}

We propose a framework that allows high-level dynamic supervision to guide complex limb articulation
policy learning, enabling robot to learn rapid turning responses for real-world dribbling tasks and become a DexDribbler. It shows improved performance compared to existing learning-based approaches, and shows capability to keep a even naturally unstable ball-surface system within control. The integration of feedback control within the reinforcement learning framework not only enhances specific skills learning like dribbling, but also has potential broader applications across various complex robotic tasks. 
% We believe that these contributions take us a step closer to realizing a fully autonomous and competitive robot team for soccer matches. 

However, it still has a number of limitations which we hope to explore and improve upon in future work: (1) Deeper integration between model-based and data-driven method, as discussed in \ref{kalman}. (2) Multi-task soccer player: recovering \cite{jiDribbleBotDynamicLegged2023} shooting \cite{jiHierarchicalReinforcementLearning2022} and goalkeeping \cite{huangCreatingDynamicQuadrupedal2022} skills could be integrated to create a fully autonomous soccer agent. (3) Awareness of other object or agent: Future work could incorporate more information about the environment geometry as well as realize high-level counter-play or cooperation with other agents. Ultimately, our aim is to develop robots that could potentially compete with humans in the near future. 

\newcommand\blfootnote[1]{%

  \begingroup[hidelinks]
  \renewcommand\thefootnote{}\footnote{#1}%
  \addtocounter{footnote}{-1}%
  \endgroup
}

%\addtolength{\textheight}{-12cm}   % This command serves to balance the column lengths
                                  % on the last page of the document manually. It shortens
                                  % the textheight of the last page by a suitable amount.
                                  % This command does not take effect until the next page
                                  % so it should come on the page before the last. Make
                                  % sure that you do not shorten the textheight too much.

 \begin{spacing}{0.86}
\bibliographystyle{IEEEtran}  %IEEEtran
\bibliography{refer.bib} %bib
\end{spacing}                                 

\vspace*{\fill}

\begin{center}\footnotesize{In memory of Dr. Yifan Liu} \\
\footnotesize{for her help and discussion in the early days of this project.}
\end{center}

\end{document}